\documentclass[runningheads]{llncs}
\usepackage{graphicx}

\newcommand{\citep}{\cite}
\newcommand{\citet}{\cite}

\newif\ifwithappendix
\withappendixfalse

\newif\ifappendixshown
\appendixshowntrue

\usepackage[T1]{fontenc}
\usepackage[utf8]{inputenc}
\usepackage[textsize=tiny]{todonotes}
\usepackage{graphicx}
\usepackage{booktabs}
\usepackage{latexsym}

\usepackage{footnote}
\makesavenoteenv{tabular}
\makesavenoteenv{table}
\makesavenoteenv{table*}

\newcommand\minput[1]{%
  \input{#1}%
  \ifhmode\ifnum\lastnodetype=11 \unskip\fi\fi}

\usepackage{hyperref}
\usepackage{xstring}

\newcommand{\noqa}[1]{}
\newcommand{\noqall}[1]{}

\usepackage{moresize}

\usepackage{amsmath}
\usepackage{multirow}
\usepackage{caption}
\usepackage{subcaption}
\usepackage{wrapfig}

\begin{document}

\title{CCpdf: Building a High Quality Corpus for Visually Rich Documents from Web Crawl Data}
\titlerunning{CCpdf: Building a High Quality Corpus for Visually Rich Documents\dots}

\author{Michał Turski\inst{1,2}\and
Tomasz Stanisławek\inst{1}\and\\
Karol Kaczmarek\inst{1,2} \and
Paweł Dyda\inst{1,2} \and
Filip Graliński\inst{1,2}}

\authorrunning{M. Turski et al.}

\institute{Snowflake\thanks{work done while at Applica.ai, later acquired by Snowflake}\\
\email{firstname.lastname@snowflake.com}\\
\url{http://snowflake.com} \and
Adam Mickiewicz University, Poznań, Poland\\
\email{firstname.lastname@amu.edu.pl}}

\maketitle              %
\begin{abstract}

In recent years, the field of document understanding has progressed a lot. A significant part of this progress has been possible thanks to the use of language models pretrained on large amounts of documents. However, pretraining corpora used in the domain of document understanding are single domain, monolingual, or nonpublic. Our goal in this paper is to propose an efficient pipeline for creating a big-scale, diverse, multilingual corpus of PDF files from all over the Internet using Common Crawl, as PDF files are the most canonical types of documents as considered in document understanding. We analyzed extensively all of the steps of the pipeline and proposed a solution which is a trade-off between data quality and processing time. 
We also share a CCpdf corpus in a form or an index of PDF files along with a script for downloading them, which produces a collection useful for language model pretraining. The dataset and tools published with this paper offer researchers the opportunity to develop even better multilingual language models.

\end{abstract}

\section{Introduction}

Natural Language Processing (NLP) in recent years has made significant progress thanks to using language models such as GPT-3 \cite{GPT-3} or T5 \cite{raffel2019exploring}. Usually these models are trained in a two-step process. The first part is pretraining, which utilizes a large corpus of text, and the second step is finetuning on a final task. Recent works demonstrate a considerable impact of pretraining on the final performance of a model \cite{Gururangan_2020,liu2019roberta,turc2019}. For instance, GPT-3 was pretrained on a combination of texts from Common Crawl, WebText2, two book corpora, and the English Wikipedia (499 billion tokens in total)\cite{GPT-3} while T5 was pretrained on the C4 corpus, which is 750 GB of data\cite{raffel2019exploring}.

The recent progress in document understanding (defined as ``capacity to convert a document into meaningful information''~\cite{Borchmann2021DUEED}) has been possible thanks to 2D language models such as LayoutLM \cite{Xu2020LayoutLMPO,xu2021layoutlmv}, LAMBERT \cite{Garncarek2021LAMBERTLL}, or TILT \cite{Powalski2021GoingFB}. Similarly to the models mentioned above, they also need large amounts of data for pretraining. The input to these models is a multi-modal representation of a document, e.g. tokens with their positions and images of pages.

\begin{figure}[t]
\includegraphics[width=0.8\linewidth]{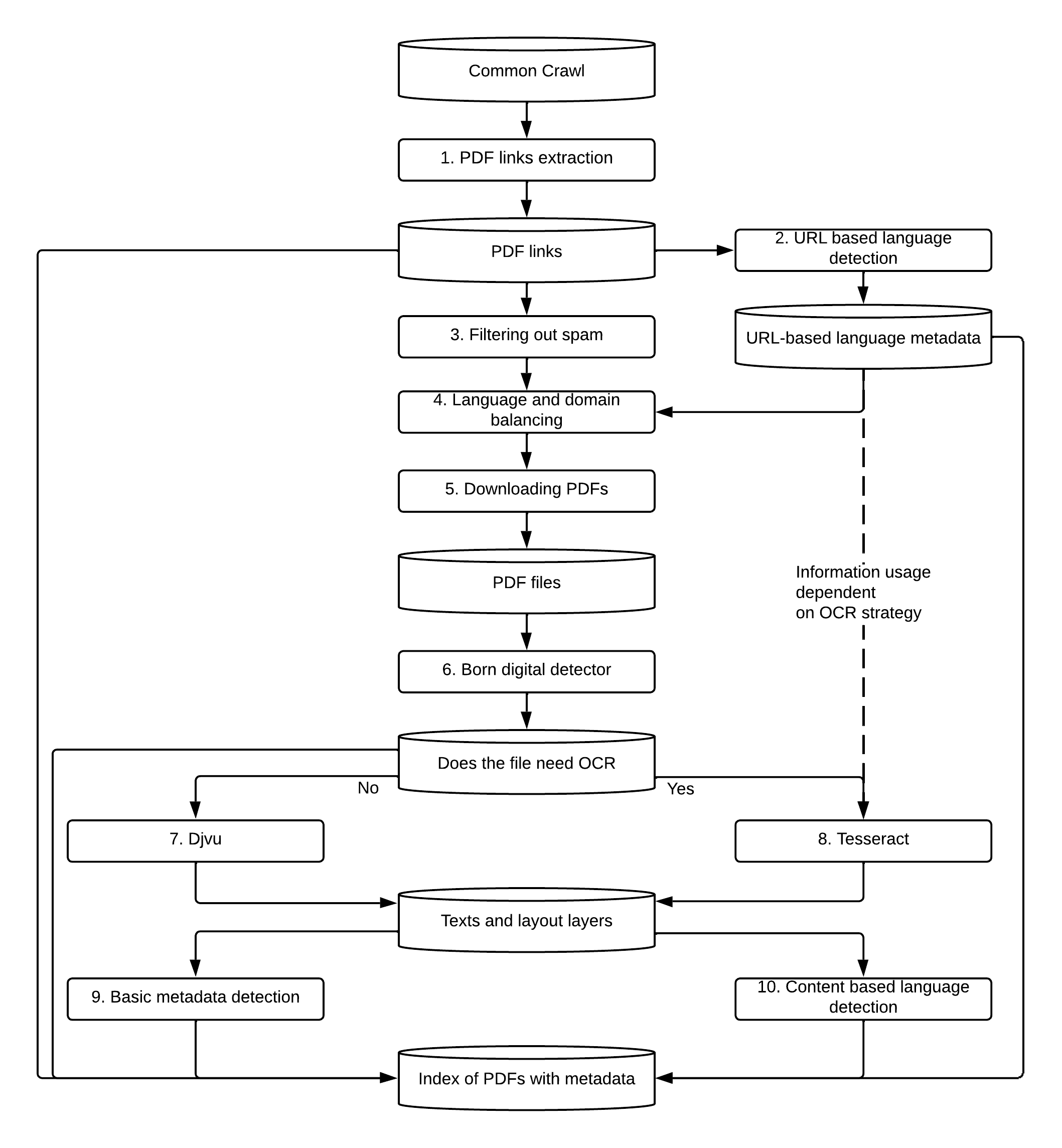}
\centering
\captionof{figure}{The full flow of the process. Cylinders represent data, rectangles represent processing steps, and arrows represent data flow. A solid line indicates that the information is always used, and a dashed line represents data usage dependent on the processing strategy.}
\label{fig:hero}
\vspace{-17pt}
\end{figure}

The World Wide Web abounds in multi-modal documents, which contain enormous amounts of information. This information can be used in multiple domains: NLP, law, knowledge extraction, history, and many more. Yet, this aspect of the Internet remains relatively unexplored. So far, attempts of document dataset creation have been focused on either single domain (e.g. medical\footnote{\scriptsize\url{www.ncbi.nlm.nih.gov/pmc/tools/openftlist/}}, academic~\cite{ammar-etal-2018-construction}, or industrial~\cite{lewis2006}), while there has been no all-over-the-Internet approach. On the other hand, existing all-over-the Internet corpora (e.g. Web 1T 5-gram\footnote{\scriptsize \url{https://catalog.ldc.upenn.edu/LDC2006T13}}) were focused on text only, not on multi-modal documents.

A document is a multi-modal form of communication: to interpret documents properly, we have to understand not only text, but also the layout and graphical elements. The most popular and portable multi-modal document format is PDF. In this study, we aim to describe a carefully designed pipeline for PDF corpus creation. We investigated numerous possible processing techniques and described their impact on the final data, which allowed us to achieve satisfactory trade-off between data quality, computing time, and monetary cost. The dataset itself (in the form of an index of PDF files and a script for downloading them) is also available at our website\footnote{\scriptsize \url{https://github.com/applicaai/CCpdf}}. We share a corpus of 14.5M pages. It is useful as a dataset for 2D language model pretraining, but may also be employed as a source for derived datasets, in the same way as the IIT-CDIP dataset~\cite{lewis2006} was used to create many diverse challenges. Finally, analysis of the collected PDF files themselves yields helpful insight for language model creators, but also enhances our understanding of the World Wide Web as a source of PDF documents.

\section{Related works}

The general problem of creating a large-scale corpus of documents has been studied extensively in recent years.
IIT-CDIP~\cite{lewis2006} is a 40M pages (but according to the authors of OCR-IDL~\cite{biten2022ocr} only 35.5M of them are still reachable) dataset of reports from the Legacy Tobacco Documents Library\footnote{\scriptsize \url{https://industrydocuments.ucsf.edu/tobacco/}} collection, which was later reused to prepare a 400k page document classification dataset \cite{harley2015icdar}.
Also, OCR-IDL~\cite{biten2022ocr} reused IIT-CDIP to publish a 26M page dataset with high-quality OCR output.
{D}o{R}e~\cite{masson-paroubek-2020-nlp} is a French dataset of 2350 annual reports from 336 companies, unfortunately the data weren't shared publicly.
There are also two layout analysis datasets based on scientific articles: Docbank~\cite{li2020docbank} and Publaynet~\cite{zhong2019publaynet}. Their volumes are 500k and 360k pages, respectively. In addition, Ammar~\textit{et~al.}~\cite{ammar-etal-2018-construction} provided corpora of scientific documents together with a literature graph (defined as ``a directed property graph which summarizes key
information in the literature and can be used to answer the queries mentioned earlier as well as more complex queries''\cite{ammar-etal-2018-construction}).
The National Library of Medicine has shared a PMC Open Access Subset\footnote{\scriptsize \url{https://ncbi.nlm.nih.gov/pmc/tools/openftlist/}} which is a corpus of open-access, open-licensed medical publications.
Allison~\textit{et~al.}~\cite{Allison_2020} proposed a pipeline for creating a corpus of PDFs sourced from the Internet. The goal of this work is to ``identify key edge cases or common deviations from the format’s specification''. They also provide analyses of files in their corpus.
All of these datasets are single-domain or single-language collections (usually both), while our aim is to create a diverse, multilingual dataset. There exists only one publication presenting such a dataset~\cite{xlm}, but the authors limited themselves to describing the data processing pipeline without analyzing their decisions. Also, their dataset was not shared.

\looseness=-1
Attempts have also been undertaken to create diversified corpora of texts sourced from the Internet. For instance, in CCNet~\cite{wenzek2019ccnet}, Common Crawl was used to create curated monolingual corpora in more than 100 languages. Also Schwenk~\textit{et~al.} used Common Crawl in CCMatrix~\cite{Schwenk2021CCMatrixMB}, but their purpose was to extract parallel sentences in different languages. The result was 10.8 billion parallel sentences in 90 languages. Another study in this vein is Smith~\textit{et~al.} \cite{smith-etal-2013-dirt}, whose method allowed to extract a 278 million token corpus of parallel English-French, English-Spanish, and English-German texts. In CCQA~\cite{huber-etal-2021-ccqa}, a method for composing multilingual question-answering task using Common Crawl was proposed. The authors shared 130 million question-answer pairs.
Liu and Curran~\cite{liu-curran-2006-web} used Open Directory Project\footnote{\scriptsize \url{http://odp.org}} to extract a topic-diverse English corpus of 10 billion words.
To pretrain the T5 language model~\cite{raffel2019exploring}, the authors extracted a 750 GB English text corpus, called C4, employing Common Crawl.
Dodge~\textit{et~al.}~\cite{dodge-etal-2021-documenting} explored this dataset further and analyzed the effects of the applied filtering.
A similar pipeline to that used for C4 was applied to create the mT5~\cite{Xue2021mT5AM} training corpus, which is a multilingual version of T5. The proposed corpus has 6.3 trillion tokens.
Qi~\textit{et~al.}~\cite{qi2020imagebert} crawled 10 million images with captions from the Internet and used it to pretrain the multi-modal ImageBERT model.
C4Corpus~\cite{habernal-etal-2016-c4corpus} (not to be confused with C4 proposed by Raffel~\textit{et~al.}, described above) utilized Common Crawl resources to provide multilingual (more than 50 languages) over 10 billion token corpus to the community.
The Pile~\cite{pile} is a 885 GB text corpus composed of 22 different datasets, and one of its subparts are texts from Common Crawl.
Abadji~\textit{et~al.}~\cite{Abadji2022TowardsAC} proposed a document-oriented multilingual 12 GB corpus of texts from Common Crawl with quality annotations. It must be noted that the authors define the term ``document'' as a long, coherent piece of text, not as a PDF file, as we do in this study.
Luccioni and Viviano~\cite{Luccioni2021WhatsIT} analyzed Common Crawl in terms of undesirable content, including hate speech and sexually explicit content, and investigated different filtering methods.

\begin{table*}[h!]
  \centering
  \ssmall
\setlength{\tabcolsep}{3pt}
\begin{tabular}{ lrrrrrr }
 \hline
 \textbf{Dataset} & \textbf{Documents} & \textbf{Pages} & \multirow{2}{1.5cm}{\raggedleft \textbf{Avg pages per doc}} & \textbf{Languages} & \textbf{Domains} & \textbf{Years} \\
 & & \\
 \hline
 IIT-CDIP & 6.5M* & 35.5M* & 5.5 & 1 & Industry documents & 1990s \\
 OCR-IDL & 4.6M &  26M & 5.7 & 1 & Industry documents & 1990s \\
 CCpdf & 1.1M & 14.5M & 12.9 & 11 & Multi-domain & Mostly 2010--2022 \\
 \hline
\end{tabular}

\caption{Comparison of existing publicly available corpora. *Numbers of valid documents/pages according to the authors of OCR-IDL~\cite{biten2022ocr}.}
\label{table:comparison}
\vspace{-15pt}
\end{table*}

\section{Collecting and processing PDFs}
In this section we describe how we addressed the challenge of finding, downloading, and processing a great volume of PDF documents. The full process is presented in Figure~\ref{fig:hero}.

\subsection{Common Crawl}

As our input we used web indexes created by Common Crawl\footnote{\scriptsize \url{https://commoncrawl.org}}. Common Crawl is a project of The Internet Archive\footnote{\scriptsize \url{https://archive.org/}} -- an organization dedicated to providing a copy of the Internet to the community. They crawl webpages and save them into crawls dumps. A crawl dump contains billions of webpages (hundreds of terabytes of uncompressed data) and a new dump has been published nearly every month since March 2014. Some earlier, more irregular dumps starting from 2008 are also available.\footnote{\scriptsize \url{https://commoncrawl.org/the-data/get-started/}} Each dump also contains an index of the crawled pages.

We decided to simply use the latest (and the largest) dump available at the time of
writing this paper — the May 2022 dump.%
\footnote{\scriptsize \url{https://commoncrawl.org/2022/06/may-2022-crawl-archive-now-available/}}
It contains 3.45 billion web pages, which amounts to 462 TB of uncompressed content. It would obviously be possible to apply the extraction procedure described in this paper to all crawls to obtain an even larger collection of PDFs, which would also allow for a diachronic analysis, but we wanted to focus on the most recent documents.

Note that dumps contain only files considered as text files by the
Common Crawl web robot. Mostly these are web pages in the HTML format,
but, fortunately, PDFs are also treated as text files, being
derivative of the PostScript page description language. This is not the case with,
for instance, images, Excel files, DOCX files. Consequently, such files
cannot be amassed using the methods described in the aforementioned papers.

\subsection{PDF links extraction}

We experimented with two methods for extracting links to PDF files (step~1 in Figure~\ref{fig:hero}):

\begin{enumerate}
\item using CDX files, i.e., index server files provided by Common Crawl;
\item looking for links to PDF files in WARC, i.e., raw crawl data files.
\end{enumerate}

\looseness=-1 The first method is simpler, as CDX files are easy to download and take up only 225 GB in total. The second method might yield more links to PDF files, but:

\begin{itemize}
\item it is impossible for us to download all WARCs. Only a limited
  number of them can be processed, though still a significant number
  of PDF links can be added even if a small percentage of all WARC
  files are processed,
\item there is lower probability that the file linked is available at all, be it in the crawl dump or simply at the original address.
\end{itemize}

In CDX files, the MIME type of a captured file is specified, and we
limited ourselves to the \texttt{application/pdf} type.

Hence, in this paper, we focus on the first method, which allows to
speed up the whole processing pipeline.

\subsection{URL-based language detection}\label{sec:langconcidered}

We decided to limit our investigation to the following set of 11 languages:
Arabic, Dutch, English, French, German, Italian, Japanese, Polish, Portuguese, Russian, and Spanish.

When deciding whether to process a given URL, we applied a number of
simple heuristics to determine the language. For example, we assumed that PDFs
from \texttt{.pl} domains are Polish unless there is \texttt{lang=en}
inside the URL etc. Note that this is a preliminary filter; later, when the
contents have been downloaded, we do a proper language detection (see
Section~\ref{sec:langident}).

In August 2018, Common Crawl added language metadata to CDX files.%
\footnote{\scriptsize \url{https://commoncrawl.org/2018/08/august-2018-crawl-archive-now-available/}}
Unfortunately, the Compact Language Detector 2 employed there is applicable only
for plain texts or HTMLs, and only a small percentage of PDF links contained the language metadata; therefore,
it was unusable for our purposes.

This step of the pipeline is presented as block~2 in Figure~\ref{fig:hero}.

\subsection{Filtering out spam}\label{sec:spamdetection}

One of the challenges to be tackled in Web information retrieval or
when creating a massive text corpus sourced from the Web is the problem of (web)
spam and, more generally, low quality pages (step~3 in Figure~\ref{fig:hero}). Web spam is usually
related to black-hat search engine optimization, i.e., creating link farms
of web pages with automatically or semi-automatically generated
content. It turns out that PDF files found on the Internet have the
advantage of a relatively low percentage of spam, especially when
compared to HTML web pages. More generally, we believe PDF files usually contain more formal
content as most of them are business, legal, or scientific documents.

Still, some spam PDFs were found in Common Crawl dumps. Fortunately,
the way in which spammers operate is rather homogeneous. A typical telltale
of a spammy PDF was a long name composed of lower-case letters
interspersed with hyphens. A regular expression was written to detect
suspicious URLs, and if a domain happened to contain a large percentage
of such URLs, it was assumed to be spammy as a whole and totally
discarded. Thanks to this simple heuristic, in a sample of 1k documents we manually annotated (see~Section~\ref{sec:langident}) we found no spam PDFs.

\subsection{PDF data download methods}

In order to ensure diversity, we downloaded at most three PDF files from each
domain for a language in a random but reproducible manner. For English and German this number was
lowered to, respectively, one and two, as PDFs in these two languages
are much more numerous compared to others. This limitation also serves as a filter
against anomalies such as millions of PDFs coming from a single domain;
especially a spammy one, if not detected with the procedure
described in Section~\ref{sec:spamdetection}. Balancing is represented as step~4 in Figure~\ref{fig:hero}.

The files were downloaded from the original URLs (step~5 in Figure~\ref{fig:hero}). Optionally, one could extract the file from a Common Crawl dump, especially if the file is not available at the original site. We provide a script to
extract PDF files directly from the dump; fortunately, one does not need
to download the whole dump to extract a file.

There is, however, one serious issue with extracting PDFs from crawl
dumps: all files are truncated by the crawler to 1 MB. This
limit is quite high for HTML pages, but unfortunately rather low for
PDF files. This means that only small-sized PDF files can be extracted
from Common Crawl dumps; larger ones have to be downloaded from the
original sites.

The final and intermediary statistics for the files downloaded are
presented in Table~\ref{table:stats}.

\begin{table*}[h!]
  \centering
  \ssmall

\setlength{\tabcolsep}{5.4pt}
\begin{tabular}{ lrrrrrr }
 \hline
  & \textbf{URLs} & \textbf{Anti-spam} & \raggedleft \textbf{Domain} & \textbf{Language} & \textbf{Successfully} & \textbf{Successfully} \\
 &  \textbf{found} & \textbf{filtered} & \textbf{balanced} & \textbf{balanced} &  \textbf{downloaded} & \textbf{processed} \\
 \hline
 ar    &     65 395 &     65 374 &    13 142 &    13 142 &    11 710 (89.10\%) &   10 826 (82.38\%)\\
 de    &  1 661 317 &  1 659 713 &   320 978 &   200 000 &   182 607 (91.30\%) &  172 668 (86.33\%)\\
 en    & 11 515 766 & 11 501 781 &   952 776 &   200 000 &   182 071 (91.04\%) &  175 440 (87.72\%)\\
 es    &    871 843 &    871 478 &   106 143 &   106 143 &    93 163 (87.77\%) &   88 952 (83.80\%)\\
 fr    &    654 250 &    653 120 &   143 020 &   143 020 &   129 927 (90.85\%) &  121 905 (85.24\%)\\
 it    &    831 344 &    831 026 &   129 610 &   129 610 &   119 731 (92.38\%) &  114 265 (88.16\%)\\
 ja    &  1 160 543 &  1 160 410 &   151 686 &   151 686 &   139 990 (92.29\%) &  134 310 (88.54\%)\\
 nl    &    339 519 &    338 946 &    92 372 &    92 372 &    84 848 (91.85\%) &   79 720 (86.30\%)\\
 pl    &    438 770 &    438 531 &    85 635 &    85 635 &    79 668 (93.03\%) &   75 374 (88.02\%)\\
 pt    &    697 535 &    697 285 &    73 130 &    73 130 &    64 725 (88.51\%) &   61 405 (83.97\%)\\
 ru    &    628 473 &    628 061 &   105 535 &   105 535 &    91 708 (86.90\%) &   85 552 (81.07\%)\\
 \hline
all    & 18 864 755 & 18 845 725 & 2 174 027 & 1 300 273 & 1 180 148 (90.76\%) & 1 120 417 (86.17\%)\\
 \hline
\end{tabular}

\caption{Number of documents per processing step and language. Percentage values show success rates of downloading (in the downloaded column) or downloading and processing together (in the processed column). The success rate for processing a downloaded document equals 94.94\%.}
\label{table:stats}
\vspace{-22pt}
\end{table*}

\subsection{
Born Digital scanner}
\label{sec:born_digital_scanner}

To process correctly all kinds of documents in the document understanding domain we need to extract tokens from PDF files with their bounding boxes sorted properly, i.e., according to the reading order. The most common approach \cite{Xu2020LayoutLMPO,xu2021layoutlmv,Garncarek2021LAMBERTLL,Powalski2021GoingFB} is to process each PDF file with the use of some OCR engine, e.g. Tesseract \cite{tesseract}, Amazon Textract\footnote{\scriptsize \url{https://aws.amazon.com/textract/}}, Microsoft Azure Computer Vision API,\footnote{\scriptsize \url{https://docs.microsoft.com/en-us/azure/cognitive-services/computer-vision/overview-ocr}} or Google Vision API\footnote{\scriptsize \url{https://cloud.google.com/vision/docs/pdf}}. This method simplifies the processing pipeline and removes the need to understand the complicated PDF file format.

The biggest challenge in direct text and layout extraction lies in processing image content since there is no easy way to detect whether an image contains text. On the other hand, some documents lack pictures altogether; instead they contain textual information in the PDF file structure. We call them \textit{documents that do not require OCR}. From such documents text can be extracted along with bounding boxes using dedicated Python libraries, such as pdfminer.six\footnote{\scriptsize \url{https://github.com/pdfminer/pdfminer.six}}, pdfplumber,\footnote{\scriptsize \url{https://github.com/jsvine/pdfplumber}} or a DjVu-based tool\footnote{\scriptsize \url{http://jwilk.net/software/pdf2djvu}, \url{https://github.com/jwilk/ocrodjvu}}. Direct text extraction using these tools leads to the reduction of the processing time and improvement of the quality of the extracted data by preventing OCR errors. Therefore, we decided to introduce a mechanism, called the Born Digital detector, for finding these kinds of documents (step~6 in Figure~\ref{fig:hero}).

\subsection{Born Digital detection heuristics}
\label{sec:born_digital_detection_heuristics}

In order to detect documents that do not need to be processed with an OCR pipeline, we created a fast, simple heuristic-based classifier:
\begin{itemize}
    \item \textit{Visible Text Length > 100} – Visible text in the document contains more than 100 characters
    \item \textit{Hidden Text Length = 0} – There is no hidden text in the document
    \item \textit{Image Count = 0} – There are no images in the document
\end{itemize}
Used statistics (\textit{Visible Text Length}, \textit{Hidden Text Length}, \textit{Image Count}) were extracted using \textit{Digital-born PDF Scanner}~\footnote{\scriptsize \url{https://github.com/applicaai/digital-born-pdf-scanner}} tool written by us.

\looseness=-1 Our simple method was able to classify 219 documents out of 967 as born-digital files that do not require OCR. (In other words, we can skip the time-consuming OCR process for more than 1 out of 5 PDF files). To check quality of our heuristic we manually annotated the same sample of documents. The precision of the proposed method was 93.15\%. All errors (15) were caused by adding a background with logo text to the file. In the future, we can also improve that kind of cases by extracting metadata information about PDF file background as well.

\begin{table*}[h!]
  \centering
  \ssmall
\setlength{\tabcolsep}{4pt}
\begin{tabular}{ lrrrrr }
 \hline
 & \textbf{Gold standard \#} & \multicolumn{4}{c}{\textbf{Born Digital detector}} \\
\cmidrule(l){3-6}
 &  & \textbf{Precision} & \textbf{Recall} & \textbf{F1-score} & \textbf{TP + FP \#} \\
 \hline
 born digital, OCR not required &  471 (48.71\%) & 93.15 & 43.31 & 59.13 & 219 (22.65\%)\\
 born digital, OCR required      &  321 (33.20\%) & - & - & - & - \\
 scan                         &  175 (18.10\%) & - & - & - & - \\
  \hline
 \hline
 all                          &  967 & - & - & - & - \\
 \hline
\end{tabular}

\caption{Results for the Born Digital detector mechanism.}
\label{table:born_digital_detector}
\end{table*}

\subsection{OCR processing}
\label{sec:ocr_processing}

One of the initial steps of the PDF processing pipeline is the URL based language detection method (see Section~\ref{sec:langconcidered}). Information about the language of the document is needed for filtering documents for specific languages and also by the OCR tool. In the next step (see Section~\ref{sec:born_digital_detection_heuristics}), we select PDF files for processing either by the DjVu-based tool (if the file is born digital then it does not require OCR) or by Tesseract OCR~\cite{tesseract}. The result is hocr files containing extracted text with its bounding boxes. This form of data serves as the input to the subsequent processing and analyzing steps.

\begin{table*}[h!]
  \centering
  \ssmall
\setlength{\tabcolsep}{5pt}
\begin{tabular}{ lrrrrr }
 \hline
 \textbf{Strategy name} & \multicolumn{2}{c}{\textbf{Processing time (using 1 CPU)}} & \multicolumn{2}{c}{\textbf{Additional cost}} \\
 \cmidrule(l){2-3}  \cmidrule(l){4-5}
 & 1k files & in relation to fastest & Single page & 1k files \\
 \hline
 DjVu-based tool + Born-digital detector        &  5.6 h~ & 1x   & - & - \\
 Tesseract + URL based LD            & 23.7 h~ & 4x  & - & - \\
 Tesseract + Built-in LD mechanism   & 75.9 h~ & 14x & - & - \\
 MSOCR + Built-in LD mechanism       & 16.7 h* & 3x & \$0.001 & \$13 \\
 \hline
\end{tabular}

\caption{Comparison of resource utilization for different strategies of the text extraction from PDF files. *for Azure OCR we used 4 CPU (which is minimal recommendation for container in version 3.2) and multiplied the number by 4.}
\label{table:ocr_alternatives}
\vspace{-22pt}
\end{table*}

\subsubsection{Possible alternatives}

In a typical scenario of extracting text with bounding boxes from a PDF file, researchers use a custom OCR engine
\cite{Xu2020LayoutLMPO,xu2021layoutlmv,Garncarek2021LAMBERTLL,Powalski2021GoingFB}, e.g. Tesseract, Microsoft Azure Computer Vision API, Amazon Textract, or Google Vision API. However, when we want to process millions of PDF files, we need to think about the utilization of resources in the context of time and money. Additionally, contrary to previous work, the language of a PDF file that we want to process is unknown. Therefore, to choose the most economical option, we tested the following strategies:
\begin{enumerate}
    \item \textit{DjVu-based tool with a born-digital detector} -- for details, please see Section~\ref{sec:born_digital_detection_heuristics}
    \item \textit{Tesseract with URL based Language Detection (LD)} -- described at the beginning of this section
    \item \textit{Tesseract with a built-in LD mechanism} -- in this strategy, we use the Tesseract OCR \cite{tesseract} engine with a built-in language detection mechanism
    \item \textit{Azure CV API with a built-in LD mechanism} -- in this strategy we use Microsoft Azure Computer Vision API\footnote{\scriptsize \url{https://docs.microsoft.com/en-us/azure/cognitive-services/computer-vision/overview-ocr}} with a built-in language detection mechanism
\end{enumerate}

We achieved the shortest processing time (see Table~\ref{table:ocr_alternatives}) with the DjVu-based tool and a born-digital detector~(see Section~\ref{sec:born_digital_detection_heuristics}), which followed from the fact that we did not need to run any ML models. Also quality of output from the DjVu-based tool is better than from any OCR engine, because it extracts real content of a file and does not introduce any processing noise. \textit{Azure CV API} and \textit{Tesseract with URL based language detection} are the slowest OCR engines with 3-4 longer processing time.
It turns out that the slowest processing strategy is \textit{Tesseract with built-in LD mechanism}, and therefore, we will not apply it in our final pipeline.

\begin{figure*}
\centering
\begin{minipage}{.5\textwidth}
  \centering
  \captionsetup{width=.9\linewidth}
  \includegraphics[width=0.8\linewidth]{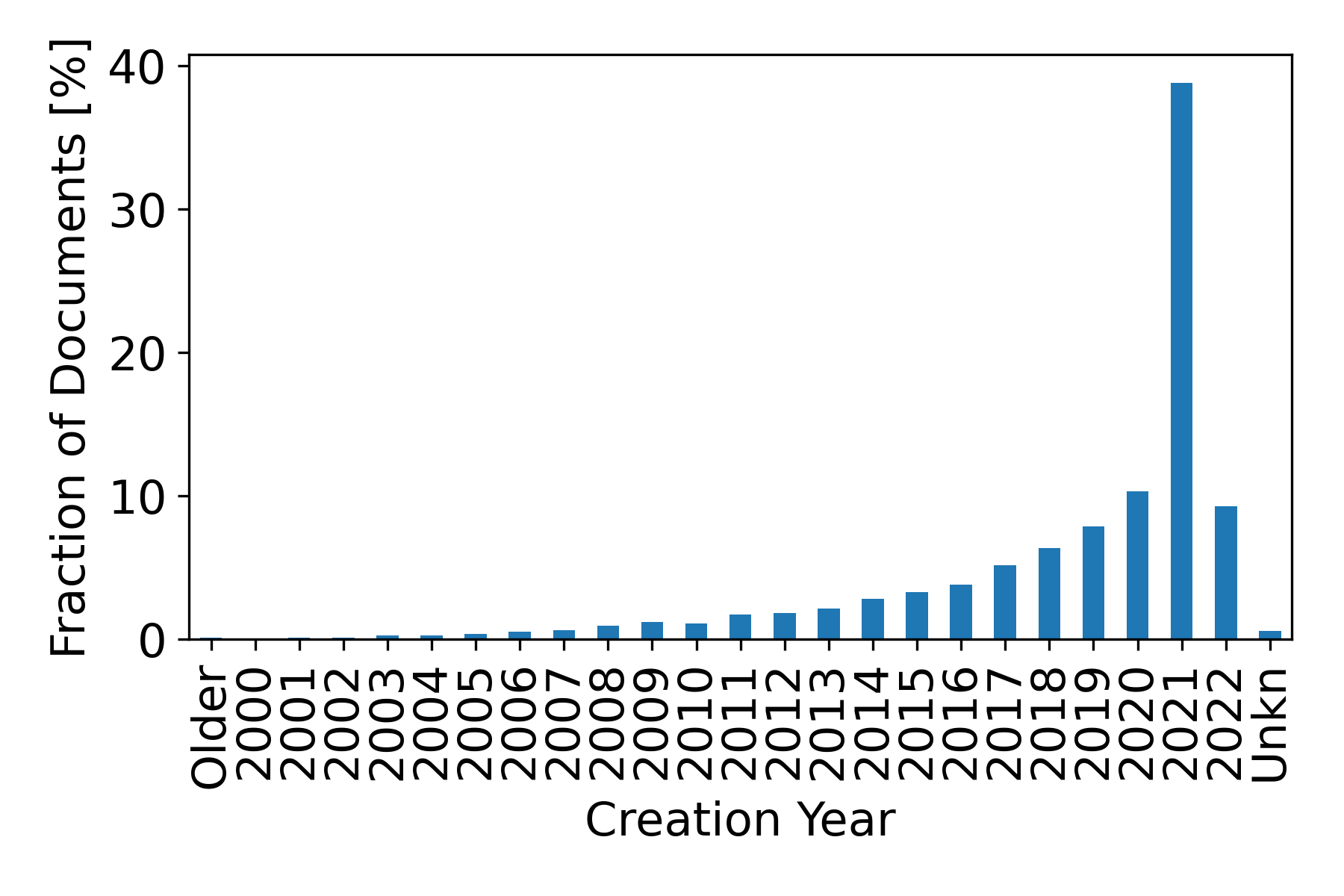}
  \captionof{figure}{Distribution of the analyzed sample in terms of creation year.}
  \label{fig:years}
\end{minipage}%
\begin{minipage}{.5\textwidth}
  \centering
  \captionsetup{width=.9\linewidth}
  \includegraphics[width=0.8\linewidth]{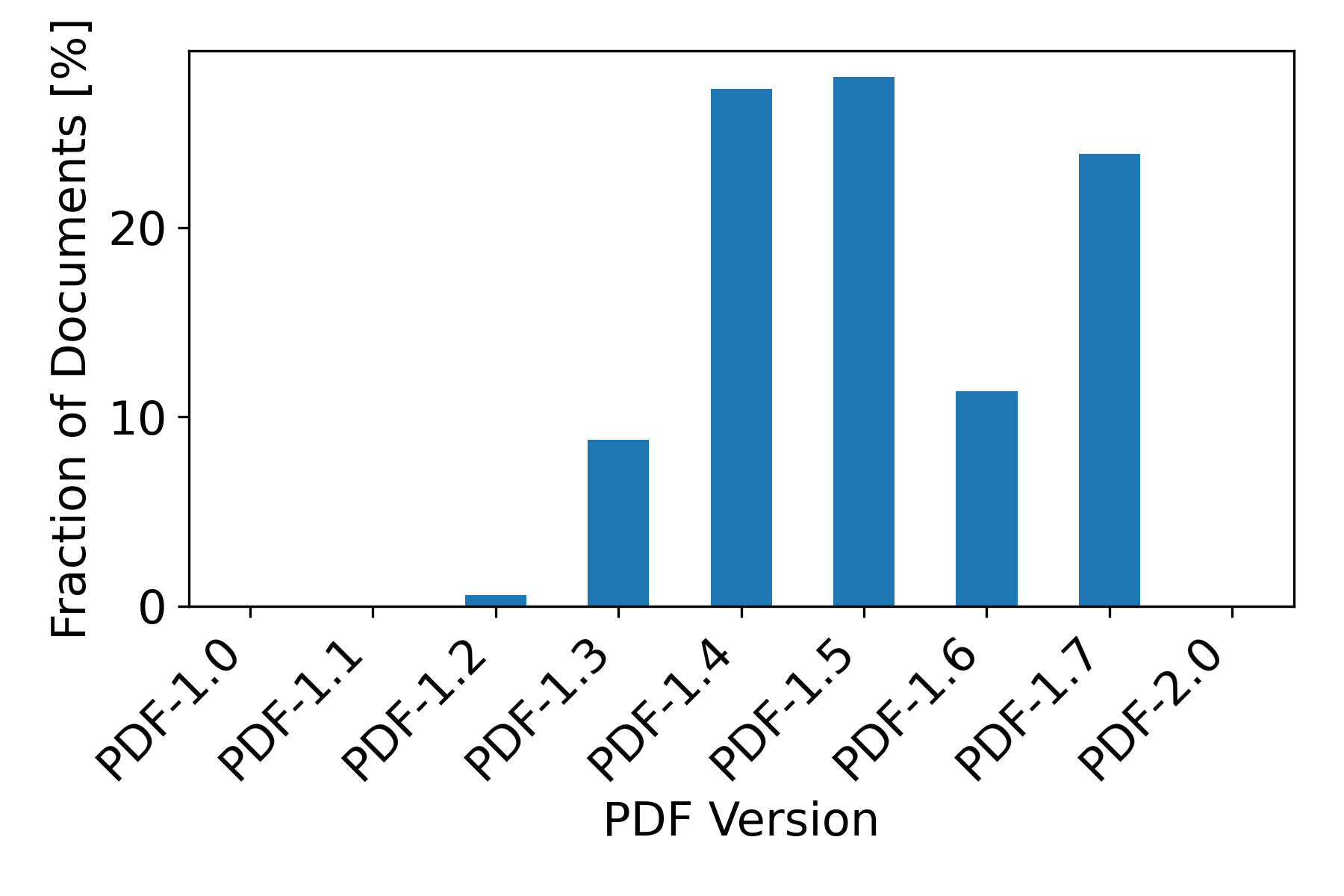}
  \captionof{figure}{Distribution of the analyzed sample according to PDF version.}
  \label{fig:pdf_version}
\end{minipage}
\end{figure*}

\subsection{Language Identification}\label{sec:langident}

In our final processing pipeline we used two language detection mechanisms:
\begin{enumerate}
    \item URL-based method. Described in Section~\ref{sec:langconcidered}.
    \item Content based method. We used the \textit{langdetect}~\footnote{\scriptsize \url{https://pypi.org/project/langdetect/}} library to detect language based on its text content extracted in the previous step.
\end{enumerate}

We tested the quality of our language detection methods on \raisebox{0.5ex}{\texttildelow}1k manually annotated documents (Table \ref{table:lang_detection}). Both of our mechanisms can detect only a single language but, in reality, we found out that 27 documents had multiple languages (in 23 cases one of them was English). Fortunately, detecting a single language allowed us to predict the language correctly for almost all documents (97.3\%).

With the use of the URL based method, we achieved a 90.51\% F1-score on average, which seems reasonably good when we take into account the simplicity of the method. The content based method works better in general with an F1-score of 94.21\%  on average. The single exception here is the Japanese language. Both mechanisms produced the least satisfactory results for two languages: Arabic and English. It turned out that many documents from the \textit{.ar} domain were actually in English. Therefore, for the content based mechanism we wrongly processed the PDF files with the Arabic Tesseract model.

Additionally, we found out that when we used the proper Tesseract model our results increased drastically to an F1-score of 98.05\% on average. The main reason why this happened was the fact that the language identification mechanism was working on the proper alphabet.

\begin{table*}[h!]
  \centering
  \ssmall

\setlength{\tabcolsep}{2.4pt}
\begin{tabular}{ lrrrrrrrrrr }
 \hline
 & \textbf{Gold} & \multicolumn{3}{c}{\textbf{URL based method}} & \multicolumn{6}{c}{\textbf{Content based method}} \\
\cmidrule(l){6-11}
 & \textbf{standard \#} & \multicolumn{3}{c}{} & \multicolumn{3}{c}{\textbf{All documents}} & \multicolumn{3}{c}{\textbf{Proper Tesseract lang}} \\
  \cmidrule(l){3-5} \cmidrule(l){6-8} \cmidrule(l){9-11}
 & & \textbf{Precision} & \textbf{Recall} & \textbf{F1} & \textbf{Precision} & \textbf{Recall} & \textbf{F1} & \textbf{Precision} & \textbf{Recall} & \textbf{F1-score} \\
 \hline
 ar    &  20 & 46.51 & 95.24 & 62.50     & 44.19 & 95.00 & 60.32   & 100.0 & 100.0 & 100.0 \\
 de    &  94 & 94.68 & 92.71 & 93.68     & 98.94 & 98.94 & 98.94   & 100.0 & 100.0 & 100.0 \\
 en    & 119 & 80.46 & 58.82 & 67.96     & 94.34 & 84.03 & 88.89   & 98.55 & 98.55 & 98.55 \\
 es    &  75 & 94.52 & 93.24 & 93.88     & 98.65 & 97.33 & 97.99   & 98.57 & 98.57 & 98.57 \\
 fr    & 108 & 93.94 & 86.92 & 90.29     & 100.0 & 91.67 & 95.65   & 100.0 & 100.0 & 100.0 \\
 it    & 101 & 93.20 & 95.05 & 94.12     & 98.97 & 95.05 & 96.97   & 94.79 & 94.79 & 94.79 \\
 jp    & 108 & 100.0 & 98.10 & 99.04     & 100.0 & 89.81 & 94.63   & 92.38 & 92.38 & 92.38 \\
 nl    &  90 & 84.91 & 100.0 & 91.84     & 98.86 & 96.67 & 97.75   & 96.67 & 96.67 & 96.67 \\
 pl    &  88 & 95.56 & 100.0 & 97.73     & 98.86 & 98.86 & 98.86   & 98.86 & 98.86 & 98.86 \\
 pt    &  83 & 94.38 & 98.82 & 96.55     & 97.62 & 98.80 & 98.21   & 98.78 & 98.78 & 98.78 \\
 ru    &  78 & 96.34 & 98.75 & 97.53     & 97.47 & 98.72 & 98.09   & 100.0 & 100.0 & 100.0 \\
 \hline
 other   &   2 & 0 & 0 & 0 & 0 & 0 & 0 & 0 & 0 & 0 \\
 \hline
 no text &   3 & 0 & 0 & 0 & 0 & 0 & 0 & 0 & 0 & 0\\
 \hline
 multi & 27 & 0 & 0 & 0 & 0 & 0 & 0 & 0 & 0 & 0\\
 \hline
 \hline
 all     & 996 & 88.59 & 92.51 & 90.51  & 93.45 & 94.99 & 94.21    & 98.05 & 98.05 & 98.05 \\
 \hline
\end{tabular}

\caption{Quality of the language identification methods verified on 996 manually annotated documents. }
\label{table:lang_detection}
\vspace{-10pt}
\end{table*}

\subsubsection{Possible alternatives}

In Table~\ref{table:lang_detection_tools} we present the results for different language identification tools. All of them achieved similar F1-scores, of which \textit{spacy} (94.33\%) and \textit{langdetect} (94.21\%) performed best. When we also take into consideration the processing time, it turns out that \textit{gcld3} was the best one with a huge advantage over the second tool, which was the \textit{langdetect} library. Therefore, we decided to balance quality and resource utilization and use \textit{langdetect} as our main tool for language identification.

\begin{table*}[h!]
  \centering
  \ssmall
\setlength{\tabcolsep}{1.6pt}
\begin{tabular}{ lrrrrrrrrrrrrrrr }
 \hline
 \textbf{Tool name} & \multicolumn{12}{c}{\textbf{F1-scores for content based method}} & \multicolumn{2}{c}{\textbf{Processing time}} \\
 \cmidrule(l){2-13} \cmidrule(l){14-15}
 & ar & de & en & es & fr & it & jp & nl & pl & pt & ru & all & 1k files & 1M files \\
 \hline
 langdetect\footnote{\scriptsize \url{https://pypi.org/project/langdetect/}}      & 60.32 & 98.94 & 88.89 & 97.99 & 95.65 & 96.97 & 94.63 & 97.75 & 98.86 & 98.21 & 98.09 & \textbf{94.21} & \textbf{0.28min} & \textbf{4.67h} \\
 lingua-py\footnote{\scriptsize \url{https://github.com/pemistahl/lingua-py}}       & 60.32 & 97.90 & 88.79 & 96.69 & 94.74 & 95.92 & 98.59 & 96.77 & 99.44 & 98.18 & 97.47 & 94.05 & 2.57min & 42.8h \\
 spacy\footnote{\scriptsize \url{https://spacy.io/universe/project/spacy_fastlang}}           & 60.32 & 98.94 & 87.33 & 97.99 & 94.79 & 97.49 & 98.59 & 97.18 & 100.0 & 98.18 & 97.47 & \textbf{94.33} & 3.62min & 60.3h \\
 gcld3\footnote{\scriptsize \url{https://pypi.org/project/gcld3/}}           & 59.01 & 98.94 & 89.91 & 97.96 & 96.15 & 97.49 & 92.16 & 97.73 & 99.44 & 98.78 & 98.07 & 94.08 & \textbf{0.03min} & \textbf{0.33h} \\
 \hline
\end{tabular}

\caption{Comparison of the quality and processing time of different language identification tools.}
\label{table:lang_detection_tools}
\vspace{-15pt}
\end{table*}

\subsection{Produced index}
As a result of our pipeline, we created an index of successfully downloaded and processed files. We decided to download up to 200k documents per language to share a reasonably sized corpus, with a good diversity of languages. It gives an acceptably good trade-off between the balance of languages and the size of the dataset. Statistics about the index are presented in Table~\ref{table:stats}.

A comparison of our dataset to existing corpora is presented in Table~\ref{table:comparison}.
The corpus we provided is smaller than the previous ones considering the total number of documents and pages. Still, language models will benefit in many aspects, (1) understanding long-distance relationships as the dataset has, on average, the longest documents compared to previous works, (2) multi-language training as we selected 11 different languages, (3) multi-domain training as we sourced documents from different websites all over the Internet, (4) document understanding of recently created documents (which may differ from the old ones in terms of language, layout, and graphical style) as the majority of files in our corpus were produced after 2010 (in IIT-CDIP, the most popular corpus so far, all the documents were created in the 90s).

\begin{figure*}
\centering
\begin{minipage}{.5\textwidth}
  \centering
  \captionsetup{width=.8\linewidth}
  \includegraphics[width=0.8\linewidth]{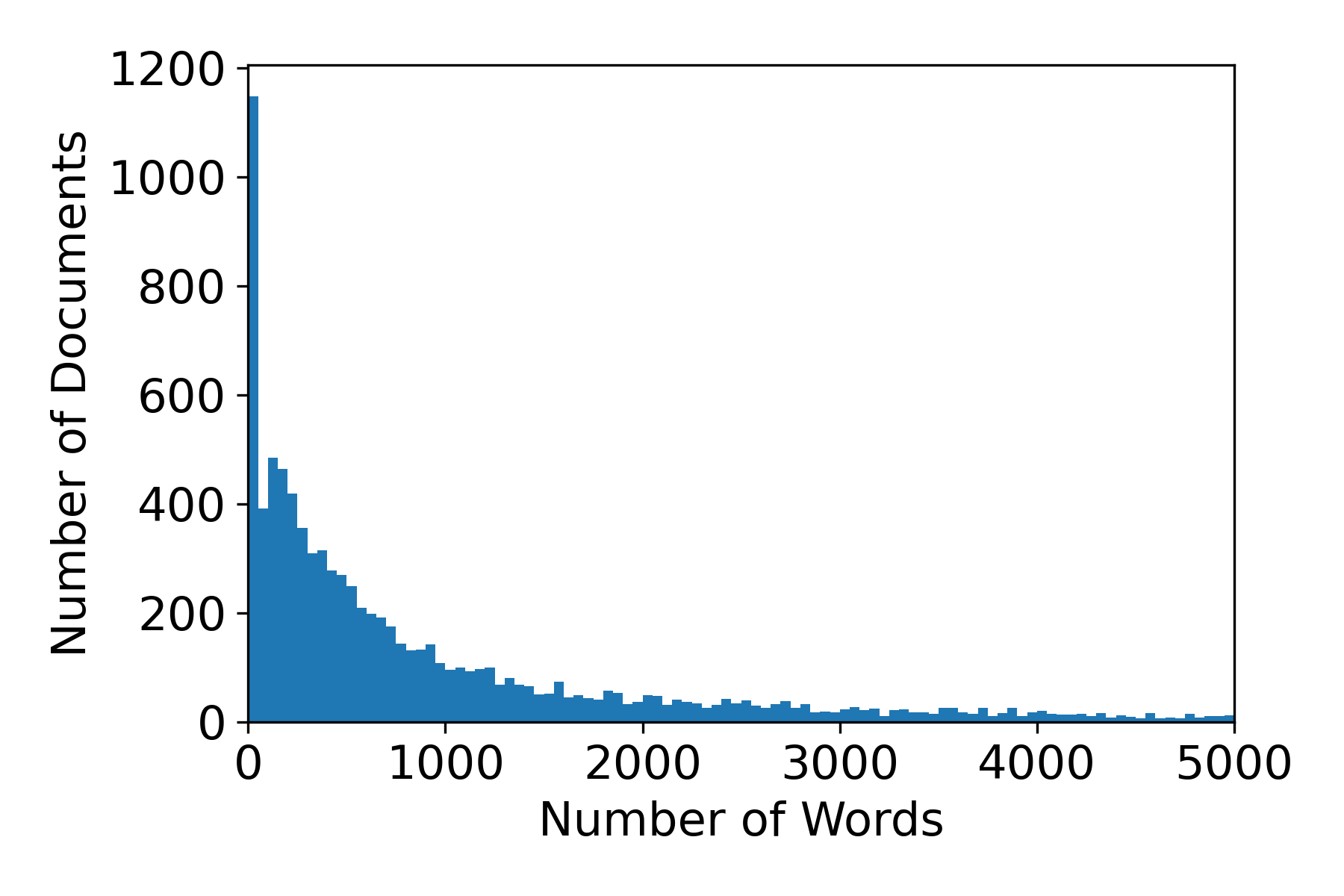}
  \captionof{figure}{Distribution of word count per document.}
  \label{fig:doc_len}
\end{minipage}%
\begin{minipage}{.5\textwidth}
  \centering
  \captionsetup{width=.8\linewidth}
  \includegraphics[width=0.8\linewidth]{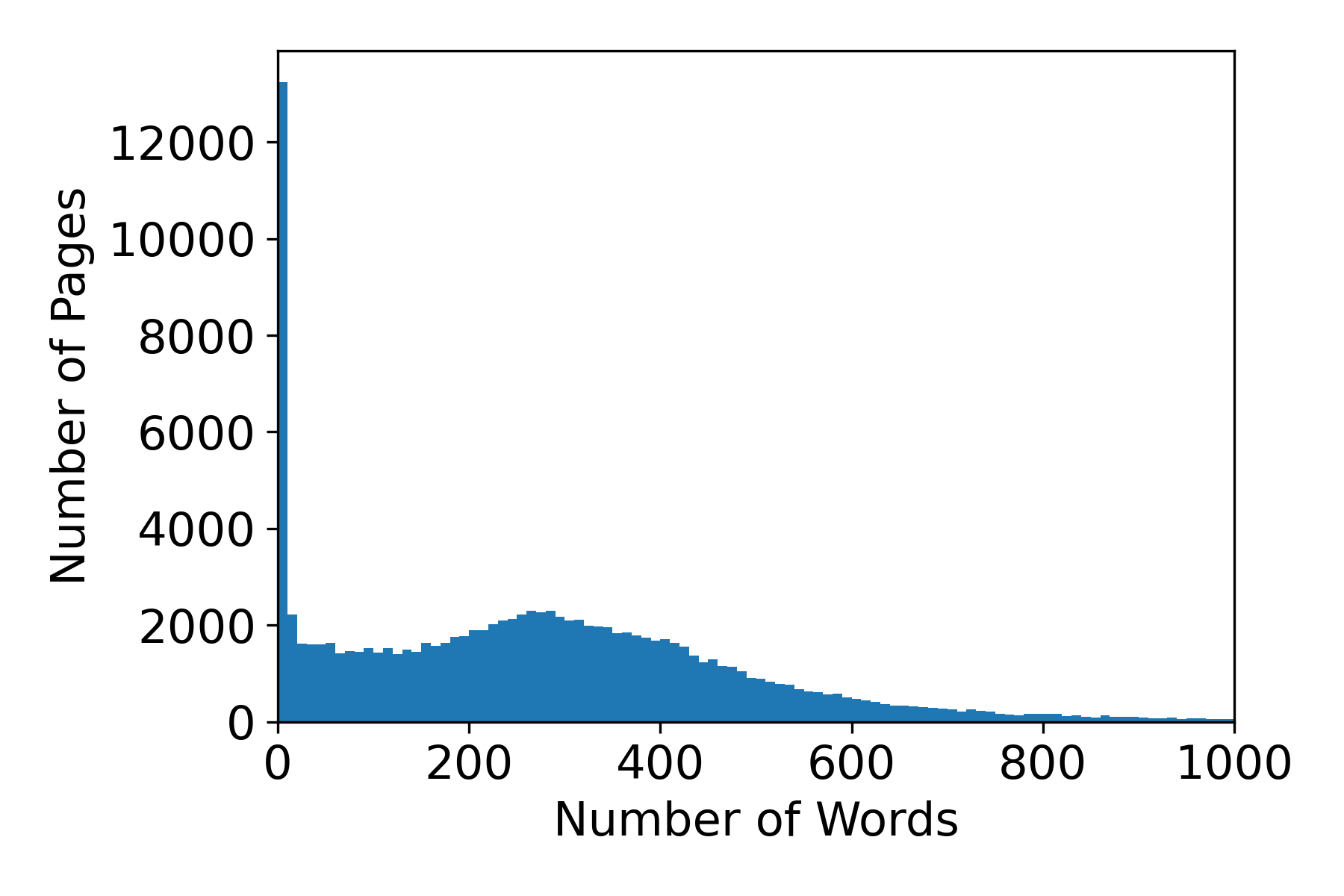}
  \captionof{figure}{Distribution of word count per page.}
  \label{fig:page_len}
\end{minipage}
\end{figure*}

\begin{figure*}
\centering
\vspace{-10pt}
\begin{minipage}{.5\textwidth}
  \centering
  \captionsetup{width=.8\linewidth}
  \includegraphics[width=0.8\linewidth]{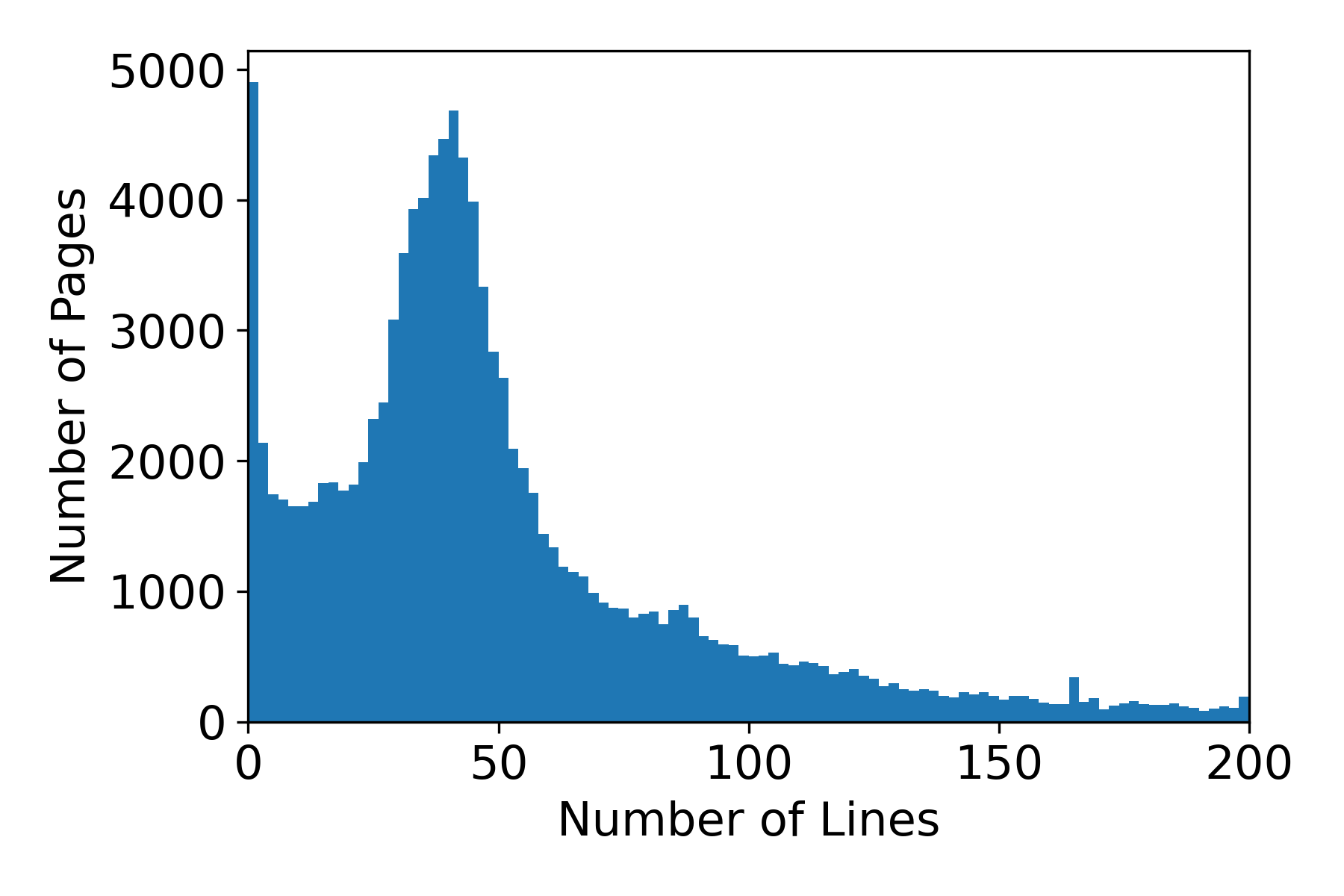}
  \captionof{figure}{Distribution of the number of lines per page.}
  \label{fig:page_lines}
\end{minipage}%
\begin{minipage}{.5\textwidth}
  \centering
  \captionsetup{width=.8\linewidth}
  \includegraphics[width=0.8\linewidth]{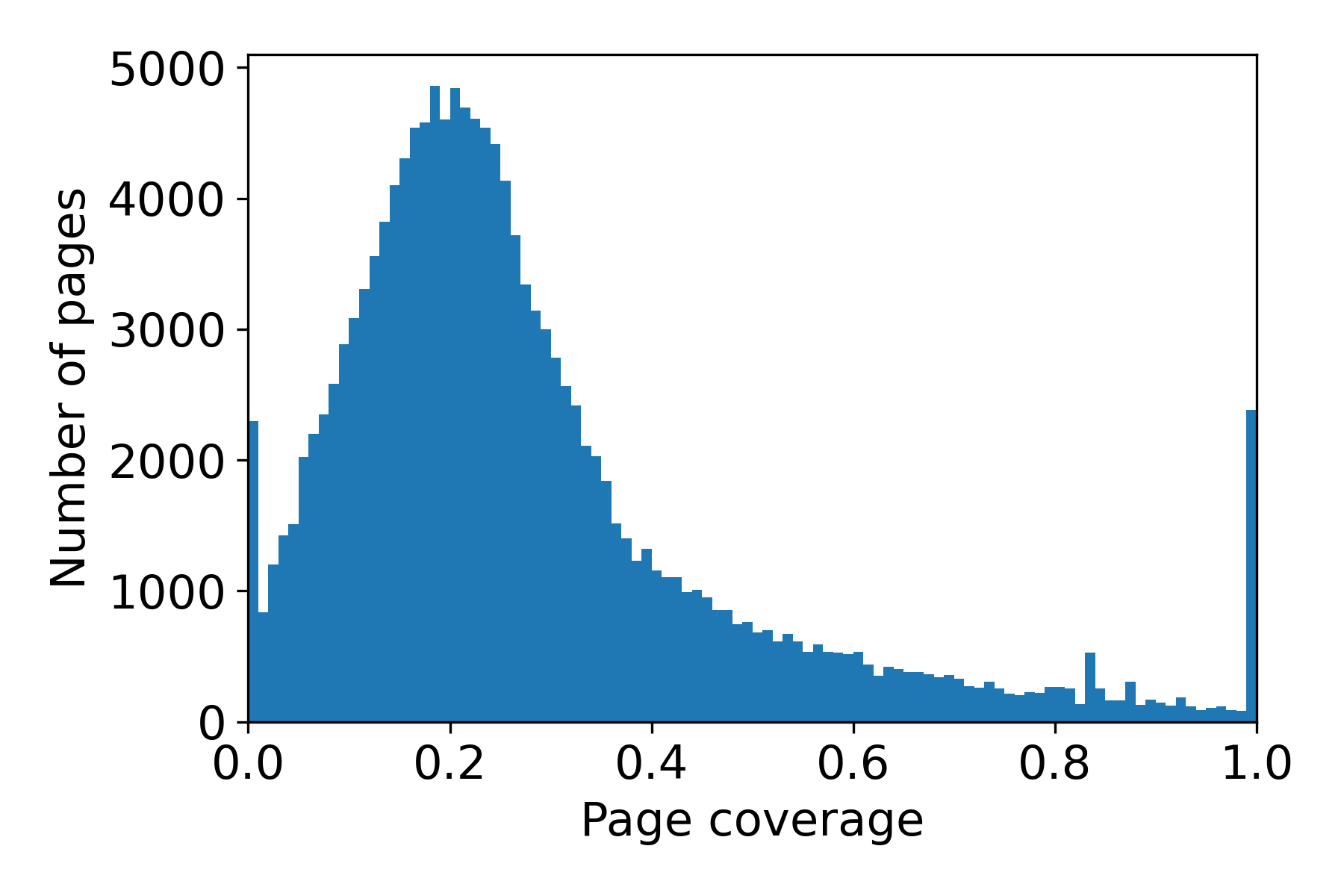}
  \captionof{figure}{Distribution of text coverage of page.}
      \label{fig:coverage}
\end{minipage}
\end{figure*}

\section{Exploration of PDFs}
Since we provide a large scale, highly diversified collection of PDFs downloaded from all over the Internet, we want to provide some insight into the properties of PDF files which are accessible on the Internet. To do so we randomly picked 1k documents in each of the languages in our corpus (11k documents at total) and analyzed them in terms of various properties.

Firstly, we analyzed them in terms of their creation date, the outcome of which is presented in Figure~\ref{fig:years}. For this analysis we used the \texttt{CreationDate} field of metadata. Since most documents come with this field filled in, we were able to read the creation date for more than 99.4\% of our sample. However, sometimes unreasonable values such as 1442 occurred as the creation year. As we can see, our corpus contains relatively new documents. It is an important point, because language evolves constantly, and three years ago terms such as ``lockdown'' or ``post-pandemic'' were absent from documents. Since we want our language models to represent current language and document types correctly, hence a distribution like that in  Figure~\ref{fig:years} is desired. The spike for the year 2021 was probably caused by the use of the Common Crawl dump from May 2022. We assume that crawlers from Common Crawl usually tend to find files that are a few months old, which often means that they are from the previous year.

We also analyzed the documents in terms of the exact version of PDF standard used. The data is presented in Figure~\ref{fig:pdf_version}. As we can see, the majority of our sample (above 76\%) are PDFs prior to the 1.7 version. It is an important property, because versions 1.7 and 2.0 are defined by an ISO standard, while the older ones were defined only by Adobe proprietary standards. Some of the issues that we experienced during processing may have been caused by problems with older standards.

\begin{wrapfigure}{l}{0.4\textwidth}
\vspace{-17pt}
\includegraphics[width=0.4\textwidth]{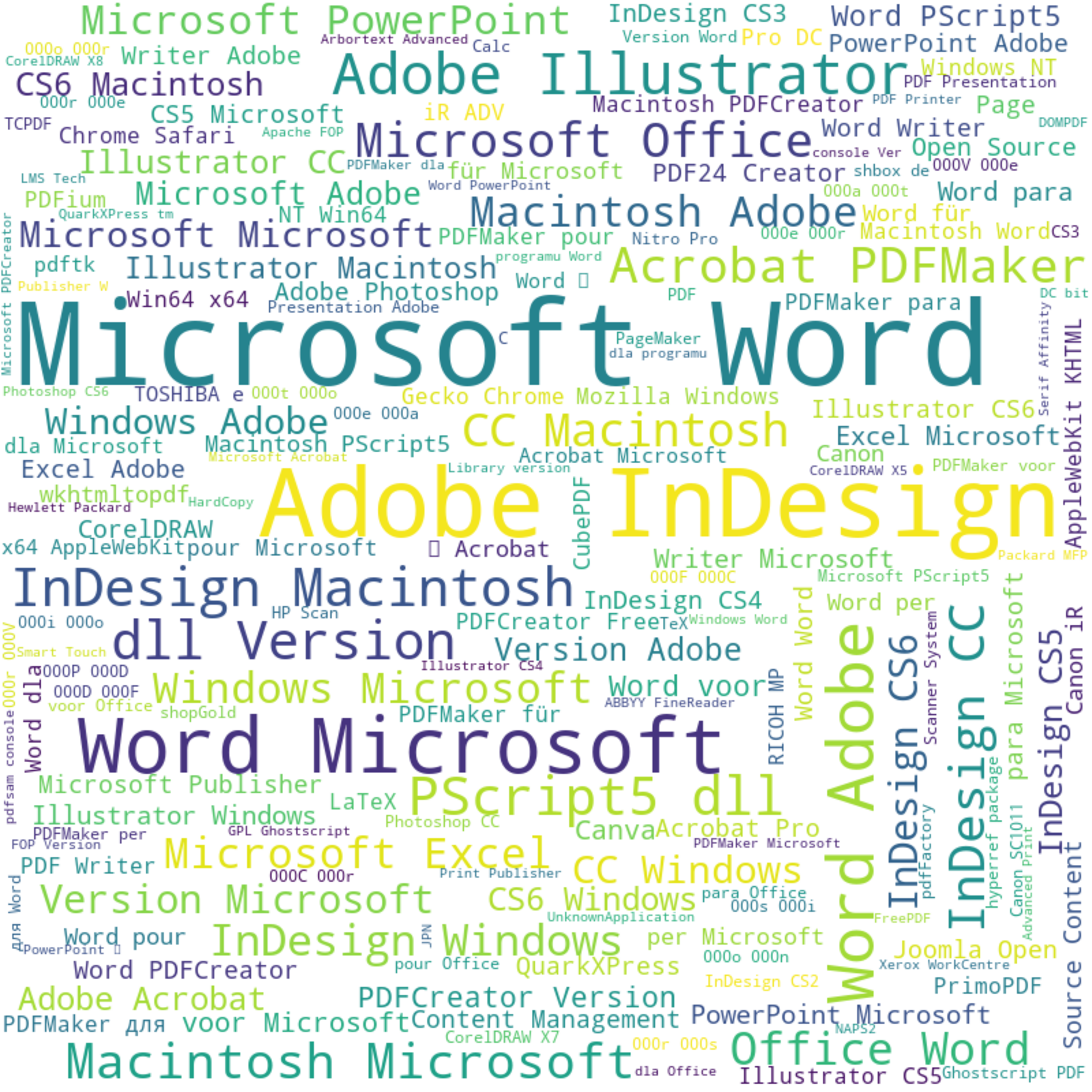} 
  \caption{Tools used to create PDFs.}
  \label{fig:creator wordcloud}
  \vspace{-22pt}
\end{wrapfigure}

A PDF file contains metadata about the tool used to create it (Figure~\ref{fig:creator wordcloud}). There are many different tools, and often the same tool was described by different values (for instance Microsoft Word has different names in many languages despite being the same program). The two most popular providers of PDF tools found in our corpus are Microsoft (29.8\% of the sample) and Adobe (21.3\% of the sample).

Other properties that we were interested in were the length of the documents in terms of word count (Figure~\ref{fig:doc_len}), their word count per page (Figure~\ref{fig:page_len}), and line count per page (Figure~\ref{fig:page_lines}). As we can see, there is great variability in terms of these parameters. For instance, there are many documents and pages with almost no text. Up to our manual check, most of the documents with little text are graphically rich, for instance, technical drawings, or infographics. The typical value of words per page is between 0 and 500, and the typical value of lines per page is between 0 and 55.

To provide some insight into the layout of the documents, we checked to what extent each page was covered by bounding boxes of tokens. We may look at it as part of the text coverage parameter. Distribution of this value is presented in Figure~\ref{fig:coverage}. 76.2\% of our sample fell into the range of 5\% to 40\% with respect to that parameter. Similarly to the previously described properties, once again we see a peak for empty pages. There is also a peak for pages fully or almost fully covered by text.

\looseness=-1 We were also interested in the ratio of page dimensions.
99.7\% of x/y ratios were in the range of 0.4 to 2; the smallest value being 0.09, and the largest -- 4.79. In our sample, 65.0\% were pages with the dimension ratio close to $\sqrt{2}$, which is a standard ratio for the A, B and C paper series. 86.9\% of them were vertical pages, and 13.1\% -- horizontal ones. Also, the LETTER format was popular; it comprised 10.6\% of the sample: 92.9\% documents were vertical, and 7.1\% horizontal. In total, the A, B, C, and LETTER series comprised 75.5\% of the sample.

\begin{figure*}
\centering
\begin{minipage}{.42\textwidth}
  \centering
  \captionsetup{width=.9\linewidth}
  \includegraphics[width=0.70\linewidth]{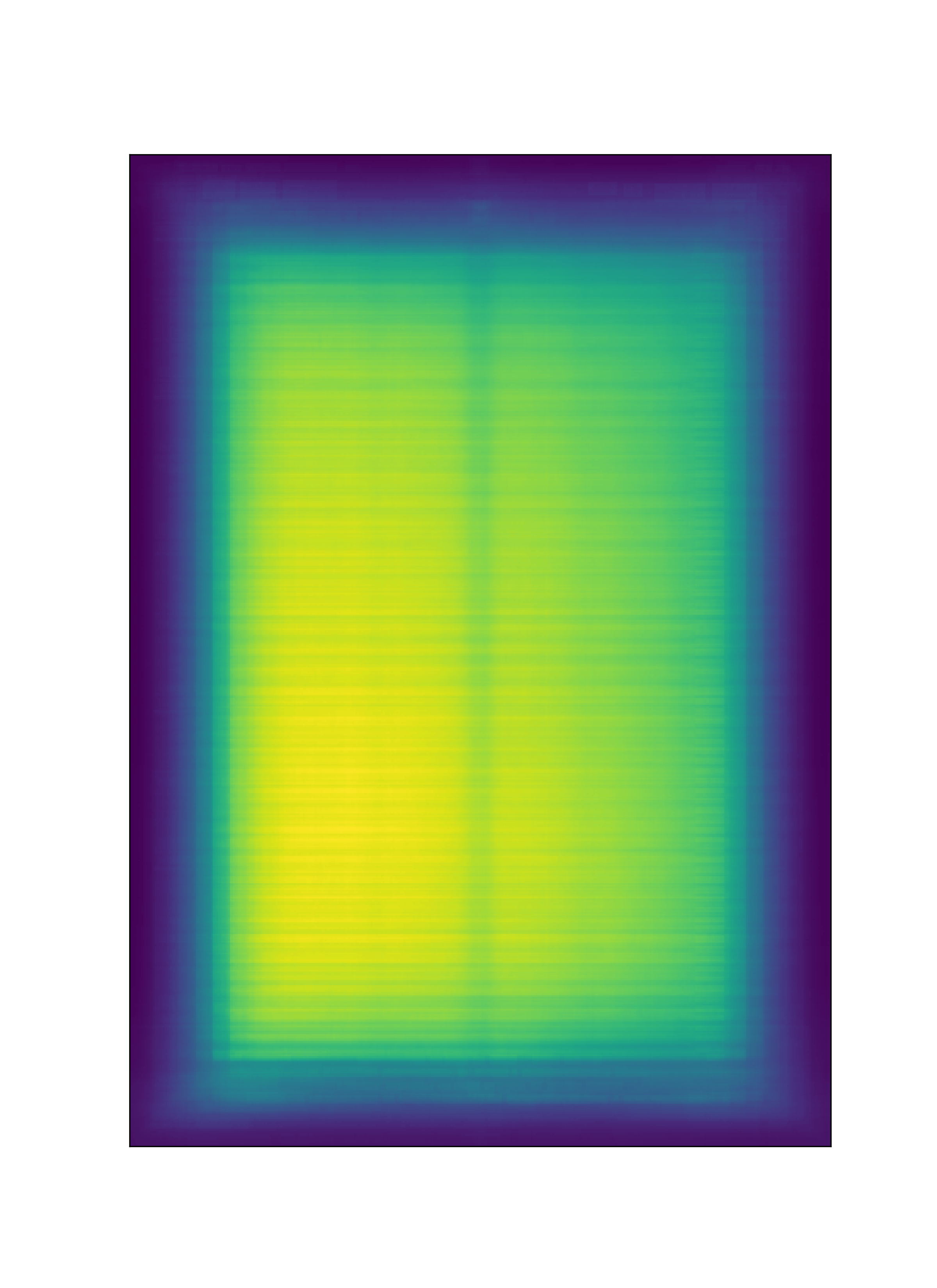}
  \captionof{figure}{Heatmap for vertical pages (brighter means more tokens, darker -- fewer tokens).}
  \label{fig:heatmap_vertical}
\end{minipage}%
\begin{minipage}{.58\textwidth}
  \centering
  \captionsetup{width=.9\linewidth}
  \includegraphics[width=0.8\linewidth]{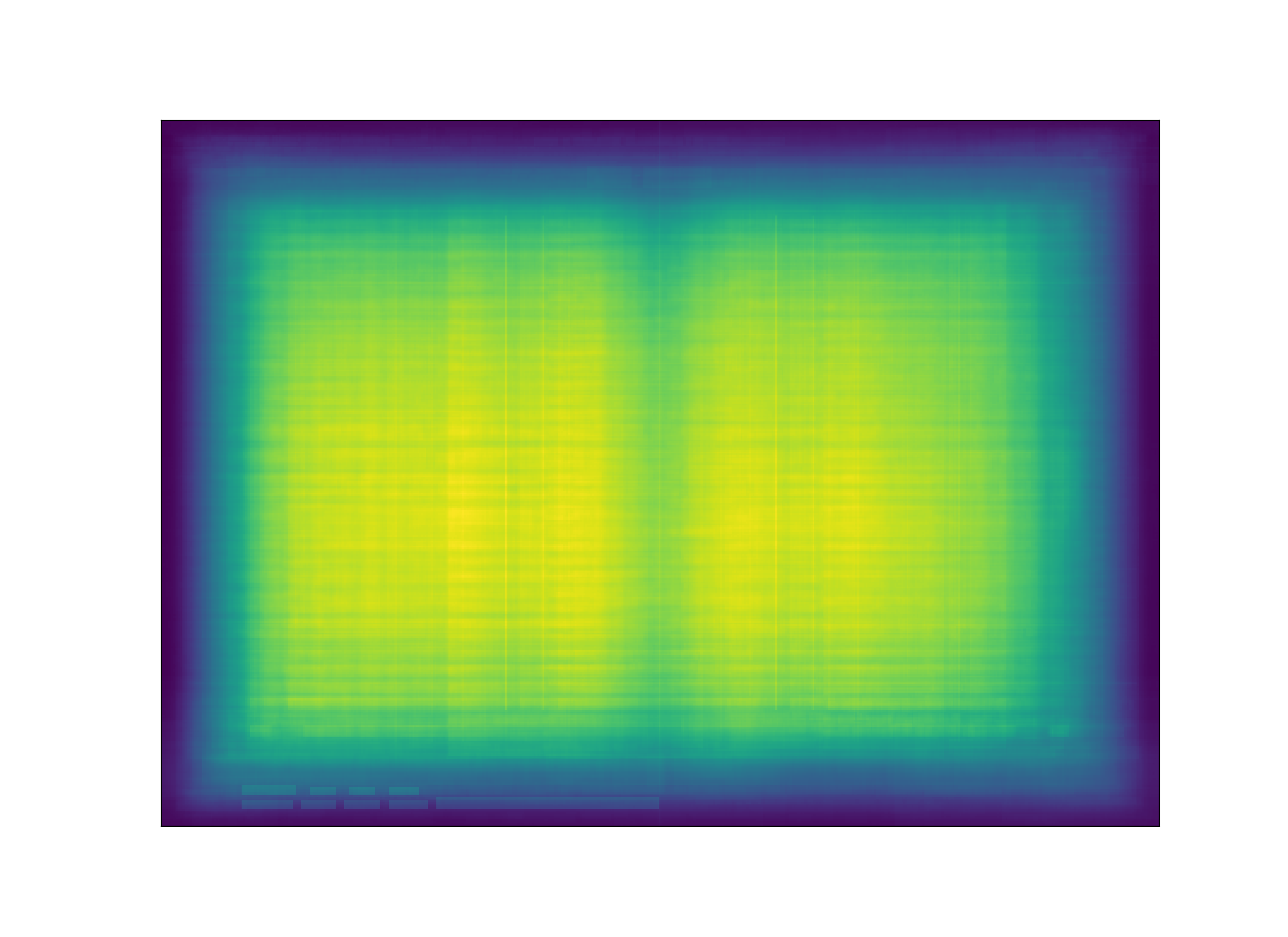}
  \captionof{figure}{Heatmap for horizontal pages (brighter means more tokens, darker -- fewer tokens).}
  \label{fig:heatmap_horizontal}

\end{minipage}
  \vspace{-12pt}
\end{figure*}

To gather more information about the layout of the documents, we created heatmaps of token bounding boxes for vertical and horizontal pages (Figures \ref{fig:heatmap_vertical} and \ref{fig:heatmap_horizontal}, respectively). As we can see, layouts with two columns of text are fairly popular, especially for horizontal pages. Also, text occurs more frequently on the left side of a page.

\section{Discussion}
In this study we analyzed a pipeline for creating a corpus of documents. According to our experiments, the most effective way of OCR processing of PDF files is a two-step procedure. The first step consists in the classification of the files according to whether they need an OCR engine or simple text extraction is sufficient. In the second step, we process the file with either an OCR engine (in our case {Tesseract}) or an extraction tool (in our case the DjVu-based tool). In the former scenario, we also discovered that predefining the OCR language speed up the process substantially; unfortunately, this comes at some cost in terms of data quality. However, this cost may be mitigated by a simple heuristic which filters out documents where the predefined OCR language did not match the one discovered by the language detector. We also analyzed different language detection tools in terms of output quality and processing time, and discovered that the \textit{langdetect} tool offered the best trade-off between these values.

One of the limitations of this research study was that we focused only on the processing pipeline without analyzing the impact of each project decision on the final language model. However, this kind of study would be very expensive, as it would require multiple pretrainings of a language model. Language model pretraining is a costly process in terms of money, time, and environmental impact \cite{Patterson2021CarbonEA}.

Also, conclusions drawn from the analysis of our sample can hardly be generalized to the whole content of the Internet and only provide some insight, rather undisputed knowledge. This follows from the filtering procedure: we decided to down-sample document-rich domains and languages, therefore, statistics calculated on the whole content of the Internet may differ from the ones presented in this work.

The approach which we used to create the dataset may be reused to all of the previous Common Crawl dumps in the WARC format, of which there are 84 in total. We decided to limit ourselves to one dump only due to computational and storage limitations. One with enough computing resources may easily reproduce our pipeline and create a corpus up to 84 times larger.

\section{Conclusions}
Large corpora of documents are crucial for 2D language model pretraining. Recent approaches to their creation have had limitations in terms of diversity and multilinguality. Diversity of the dataset is a crucial property, as data used in the training phase impact the biases of the model. Efficient design of a pipeline for creating such a corpus has not been studied before. In this work we addressed those limitations by designing a process of downloading diversified samples of PDFs and their efficient processing. To obtain documents we used Common Crawl, which is a popular source of data for language model pretraining, but has rarely been used in the context of 2D language models. The PDF files used for this project were balanced across languages and domains, which guarantees diversity with respect to layouts and topics. To make the processing pipeline efficient in terms of computing time and data quality, we tested different strategies of OCR processing, i.e. usage of the embedded textual layer for documents not requiring OCR, and predefining the OCR language. The language detection step was also carefully analyzed.

The result of this work is an index of PDF files with their URL addresses and metadata, and the script for downloading it is available at our repository\footnote{\scriptsize \url{https://github.com/applicaai/CCpdf}}. The supplied data were analyzed in terms of not only document length and layout, but also metadata connected to the PDF format (i.e., the PDF version and the creator tool), which can help understand better the dataset itself, but also give an insight into the content of the Internet.

\section*{Acknowledgments}
    
The Smart Growth Operational Programme partially supported this research under projects no. POIR.01.01.01-00-0877/19-00 (\textit{A universal platform for robotic automation of processes requiring text comprehension, with a unique level of implementation and service automation}) and POIR.01.01.01-00-1624/20 (\textit{Hiper-OCR - an innovative solution for information extraction from scanned documents}).

\bibliography{ms}

\begin{thebibliography}{10}
\providecommand{\url}[1]{\texttt{#1}}
\providecommand{\urlprefix}{URL }
\providecommand{\doi}[1]{https://doi.org/#1}

\bibitem{Abadji2022TowardsAC}
Abadji, J., Suarez, P.O., Romary, L., Sagot, B.: Towards a cleaner
  document-oriented multilingual crawled corpus. ArXiv abs/2201.06642  (2022)

\bibitem{Allison_2020}
Allison, T., Burke, W., Constantinou, V., Goh, E., Mattmann, C., Mensikova, A.,
  Southam, P., Stonebraker, R., Timmaraju, V.: Research report: Building a wide
  reach corpus for secure parser development. In: 2020 IEEE Security and
  Privacy Workshops (SPW). pp. 318--326 (2020).
  \doi{10.1109/SPW50608.2020.00066}

\bibitem{ammar-etal-2018-construction}
Ammar, W., Groeneveld, D., Bhagavatula, C., Beltagy, I., Crawford, M., Downey,
  D., Dunkelberger, J., Elgohary, A., Feldman, S., Ha, V., Kinney, R.,
  Kohlmeier, S., Lo, K., Murray, T., Ooi, H.H., Peters, M., Power, J.,
  Skjonsberg, S., Wang, L.L., Wilhelm, C., Yuan, Z., van Zuylen, M., Etzioni,
  O.: Construction of the literature graph in semantic scholar. In: Proceedings
  of the 2018 Conference of the North {A}merican Chapter of the Association for
  Computational Linguistics: Human Language Technologies, Volume 3 (Industry
  Papers). pp. 84--91. Association for Computational Linguistics, New Orleans -
  Louisiana (Jun 2018). \doi{10.18653/v1/N18-3011},
  \url{https://aclanthology.org/N18-3011}

\bibitem{biten2022ocr}
Biten, A.F., Tito, R., Gomez, L., Valveny, E., Karatzas, D.: {OCR-IDR}: {OCR}
  annotations for industry document library dataset. arXiv preprint
  arXiv:2202.12985  (2022)

\bibitem{Borchmann2021DUEED}
Borchmann, {\L}., Pietruszka, M., Stanisławek, T., Jurkiewicz, D., Turski, M.,
  Szyndler, K., Gralinski, F.: {DUE}: {End}-to-end document understanding
  benchmark. In: NeurIPS Datasets and Benchmarks (2021)

\bibitem{GPT-3}
Brown, T., Mann, B., Ryder, N., Subbiah, M., Kaplan, J.D., Dhariwal, P.,
  Neelakantan, A., Shyam, P., Sastry, G., Askell, A., Agarwal, S.,
  Herbert-Voss, A., Krueger, G., Henighan, T., Child, R., Ramesh, A., Ziegler,
  D., Wu, J., Winter, C., Hesse, C., Chen, M., Sigler, E., Litwin, M., Gray,
  S., Chess, B., Clark, J., Berner, C., McCandlish, S., Radford, A., Sutskever,
  I., Amodei, D.: Language models are few-shot learners. In: Larochelle, H.,
  Ranzato, M., Hadsell, R., Balcan, M., Lin, H. (eds.) Advances in Neural
  Information Processing Systems. vol.~33, pp. 1877--1901. Curran Associates,
  Inc. (2020),
  \url{https://proceedings.neurips.cc/paper/2020/file/1457c0d6bfcb4967418bfb8ac142f64a-Paper.pdf}

\bibitem{dodge-etal-2021-documenting}
Dodge, J., Sap, M., Marasovi{\'c}, A., Agnew, W., Ilharco, G., Groeneveld, D.,
  Mitchell, M., Gardner, M.: Documenting large webtext corpora: A case study on
  the colossal clean crawled corpus. In: Proceedings of the 2021 Conference on
  Empirical Methods in Natural Language Processing. pp. 1286--1305. Association
  for Computational Linguistics, Online and Punta Cana, Dominican Republic (Nov
  2021). \doi{10.18653/v1/2021.emnlp-main.98},
  \url{https://aclanthology.org/2021.emnlp-main.98}

\bibitem{pile}
Gao, L., Biderman, S., Black, S., Golding, L., Hoppe, T., Foster, C., Phang,
  J., He, H., Thite, A., Nabeshima, N., Presser, S., Leahy, C.: The {P}ile: An
  800gb dataset of diverse text for language modeling. arXiv preprint
  arXiv:2101.00027  (2020)

\bibitem{Garncarek2021LAMBERTLL}
Garncarek, {\L}., Powalski, R., Stanis{\l}awek, T., Topolski, B., Halama, P.,
  Turski, M., Grali{\'{n}}ski, F.: {LAMBERT}: {L}ayout-aware language modeling
  for information extraction. In: Llad{\'o}s, J., Lopresti, D., Uchida, S.
  (eds.) Document Analysis and Recognition -- ICDAR 2021. pp. 532--547.
  Springer International Publishing, Cham (2021).
  \doi{10.1007/978-3-030-86549-8_34},
  \url{https://link.springer.com/content/pdf/10.1007%2F978-3-030-86549-8_34.pdf}

\bibitem{Gururangan_2020}
Gururangan, S., Marasović, A., Swayamdipta, S., Lo, K., Beltagy, I., Downey,
  D., Smith, N.A.: Don’t stop pretraining: Adapt language models to domains
  and tasks. Proceedings of the 58th Annual Meeting of the Association for
  Computational Linguistics  (2020). \doi{10.18653/v1/2020.acl-main.740},
  \url{http://dx.doi.org/10.18653/v1/2020.acl-main.740}

\bibitem{habernal-etal-2016-c4corpus}
Habernal, I., Zayed, O., Gurevych, I.: {C}4{C}orpus: Multilingual web-size
  corpus with free license. In: Proceedings of the Tenth International
  Conference on Language Resources and Evaluation ({LREC}'16). pp. 914--922.
  European Language Resources Association (ELRA), Portoro{\v{z}}, Slovenia (May
  2016), \url{https://aclanthology.org/L16-1146}

\bibitem{harley2015icdar}
Harley, A.W., Ufkes, A., Derpanis, K.G.: Evaluation of deep convolutional nets
  for document image classification and retrieval. In: ICDAR (2015)

\bibitem{huber-etal-2021-ccqa}
Huber, P., Aghajanyan, A., Oğuz, B., Okhonko, D., tau Yih, W., Gupta, S.,
  Chen, X.: {CCQA}: {A} new web-scale question answering dataset for model
  pre-training (2021)

\bibitem{lewis2006}
Lewis, D., Agam, G., Argamon, S., Frieder, O., Grossman, D., Heard, J.:
  Building a test collection for complex document information processing. In:
  Proceedings of the 29th Annual International ACM SIGIR Conference on Research
  and Development in Information Retrieval. p. 665–666. SIGIR '06,
  Association for Computing Machinery, New York, NY, USA (2006).
  \doi{10.1145/1148170.1148307}, \url{https://doi.org/10.1145/1148170.1148307}

\bibitem{li2020docbank}
Li, M., Xu, Y., Cui, L., Huang, S., Wei, F., Li, Z., Zhou, M.: Docbank: A
  benchmark dataset for document layout analysis (2020)

\bibitem{liu-curran-2006-web}
Liu, V., Curran, J.R.: Web text corpus for natural language processing. In:
  11th Conference of the {E}uropean Chapter of the Association for
  Computational Linguistics. Association for Computational Linguistics, Trento,
  Italy (Apr 2006), \url{https://www.aclweb.org/anthology/E06-1030}

\bibitem{liu2019roberta}
Liu, Y., Ott, M., Goyal, N., Du, J., Joshi, M., Chen, D., Levy, O., Lewis, M.,
  Zettlemoyer, L., Stoyanov, V.: {RoBERTa}: {A} robustly optimized {BERT}
  pretraining approach (2019)

\bibitem{Luccioni2021WhatsIT}
Luccioni, A.S., Viviano, J.D.: What’s in the box? {An} analysis of
  undesirable content in the {Common Crawl} corpus. In: ACL (2021)

\bibitem{masson-paroubek-2020-nlp}
Masson, C., Paroubek, P.: {NLP} analytics in finance with {D}o{R}e: A {F}rench
  250{M} tokens corpus of corporate annual reports. In: Proceedings of The 12th
  Language Resources and Evaluation Conference. pp. 2261--2267. European
  Language Resources Association, Marseille, France (May 2020),
  \url{https://www.aclweb.org/anthology/2020.lrec-1.275}

\bibitem{Patterson2021CarbonEA}
Patterson, D.A., Gonzalez, J., Le, Q.V., Liang, C., Mungu{\'i}a, L.M.,
  Rothchild, D., So, D.R., Texier, M., Dean, J.: Carbon emissions and large
  neural network training. ArXiv abs/2104.10350  (2021)

\bibitem{Powalski2021GoingFB}
Powalski, R., Borchmann, {\L}., Jurkiewicz, D., Dwojak, T., Pietruszka, M.,
  Pa{\l}ka, G.: Going full-{TILT} boogie on document understanding with
  text-image-layout transformer. In: Llad{\'o}s, J., Lopresti, D., Uchida, S.
  (eds.) Document Analysis and Recognition -- ICDAR 2021. pp. 732--747.
  Springer International Publishing, Cham (2021).
  \doi{10.1007/978-3-030-86331-9_47},
  \url{https://link.springer.com/content/pdf/10.1007%2F978-3-030-86331-9_47.pdf}

\bibitem{qi2020imagebert}
Qi, D., Su, L., Song, J., Cui, E., Bharti, T., Sacheti, A.: Imagebert:
  Cross-modal pre-training with large-scale weak-supervised image-text data
  (2020)

\bibitem{raffel2019exploring}
Raffel, C., Shazeer, N., Roberts, A., Lee, K., Narang, S., Matena, M., Zhou,
  Y., Li, W., Liu, P.J.: Exploring the limits of transfer learning with a
  unified text-to-text transformer (2019)

\bibitem{Schwenk2021CCMatrixMB}
Schwenk, H., Wenzek, G., Edunov, S., Grave, E., Joulin, A.: {CCMatrix}:
  {Mining} billions of high-quality parallel sentences on the web. In: ACL
  (2021)

\bibitem{smith-etal-2013-dirt}
Smith, J.R., Saint-Amand, H., Plamada, M., Koehn, P., Callison-Burch, C.,
  Lopez, A.: Dirt cheap web-scale parallel text from the {Common} {Crawl}. In:
  Proceedings of the 51st Annual Meeting of the Association for Computational
  Linguistics (Volume 1: Long Papers). pp. 1374--1383. Association for
  Computational Linguistics, Sofia, Bulgaria (Aug 2013),
  \url{https://www.aclweb.org/anthology/P13-1135}

\bibitem{tesseract}
Smith, R.: {Tesseract Open Source OCR Engine} (2022),
  \url{https://github.com/tesseract-ocr/tesseract}

\bibitem{turc2019}
Turc, I., Chang, M.W., Lee, K., Toutanova, K.: Well-read students learn better:
  On the importance of pre-training compact models. arXiv preprint
  arXiv:1908.08962v2  (2019)

\bibitem{wenzek2019ccnet}
Wenzek, G., Lachaux, M.A., Conneau, A., Chaudhary, V., Guzmán, F., Joulin, A.,
  Grave, E.: {CCNet}: {Extracting} high quality monolingual datasets from web
  crawl data (2019)

\bibitem{xu2021layoutlmv}
Xu, Y., Xu, Y., Lv, T., Cui, L., Wei, F., Wang, G., Lu, Y., Florencio, D.,
  Zhang, C., Che, W., Zhang, M., Zhou, L.: {LayoutLMv2}: {Multi-modal}
  pre-training for visually-rich document understanding. In: ACL-IJCNLP 2021
  (January 2021)

\bibitem{Xu2020LayoutLMPO}
Xu, Y., Li, M., Cui, L., Huang, S., Wei, F., Zhou, M.: {LayoutLM}:
  {Pre-training} of text and layout for document image understanding. In:
  Proceedings of the 26th ACM SIGKDD International Conference on Knowledge
  Discovery \& Data Mining (2020)

\bibitem{xlm}
Xu, Y., Lv, T., Cui, L., Wang, G., Lu, Y., Florencio, D., Zhang, C., Wei, F.:
  Layoutxlm: Multimodal pre-training for multilingual visually-rich document
  understanding. arXiv preprint arXiv:2004.2104.08836  (2021)

\bibitem{Xue2021mT5AM}
Xue, L., Constant, N., Roberts, A., Kale, M., Al-Rfou, R., Siddhant, A., Barua,
  A., Raffel, C.: {mT5}: {A} massively multilingual pre-trained text-to-text
  transformer. In: NAACL (2021)

\bibitem{zhong2019publaynet}
Zhong, X., Tang, J., Yepes, A.J.: Publaynet: largest dataset ever for document
  layout analysis. In: 2019 International Conference on Document Analysis and
  Recognition (ICDAR). pp. 1015--1022. IEEE (Sep 2019).
  \doi{10.1109/ICDAR.2019.00166}

\end{thebibliography}
\bibliographystyle{splncs04}

\ifwithappendix
\input{appendix}
\fi

\end{document}